%% file: main.tex
\documentclass[10pt,twocolumn,letterpaper]{article}

\usepackage{cvpr}              

\input{preamble}

\definecolor{cvprblue}{rgb}{0.21,0.49,0.74}
\usepackage[pagebackref,breaklinks,colorlinks,allcolors=cvprblue]{hyperref}

\usepackage{xcolor}

\def\paperID{32860} 
\def\confName{CVPR}
\def\confYear{2026}

\title{VAGNet: Grounding 3D Affordance from Human-Object Interactions in Videos}

\author{Aihua Mao\textsuperscript{1} \quad Kaihang Huang\textsuperscript{1} \quad Yong-Jin Liu\textsuperscript{2} \quad Chee Seng Chan\textsuperscript{3} \quad Ying He\textsuperscript{4} \\ \\
\textsuperscript{1}SCUT \quad \textsuperscript{2}THU \quad \textsuperscript{3}UM \quad \textsuperscript{4}NTU\\
}

\begin{document}
\maketitle
\input{sec/0_abstract_new}    
\input{sec/1_introduction_new}
\input{sec/2_related}

\input{sec/3_method}

\input{sec/4_experiments}

\input{sec/5_conclusion}


{
    \clearpage  
    \small
    \bibliographystyle{ieeenat_fullname}
    \bibliography{main}
}

\end{document}

%% file: preamble.tex
\usepackage{multicol}
\usepackage{multirow}
\usepackage{graphicx}
\usepackage{amsmath}
\usepackage{longtable}
\usepackage{xcolor}
\usepackage{pifont}
\usepackage{placeins}
\usepackage{booktabs}

%% file: sec/0_abstract_new.tex
\begin{abstract}

3D object affordance grounding aims to identify regions on 3D objects that support human–object interaction (HOI), a capability essential to embodied visual reasoning. However, most existing approaches rely on static visual or textual cues, neglecting that affordances are inherently defined by dynamic actions. As a result, they often struggle to localize the true contact regions involved in real interactions. We take a different perspective. Humans learn how to use objects by observing and imitating actions, not just by examining shapes. Motivated by this intuition, we introduce video-guided 3D affordance grounding, which leverages dynamic interaction sequences to provide functional supervision. To achieve this, we propose VAGNet, a framework that aligns video-derived interaction cues with 3D structure to resolve ambiguities that static cues cannot address. To support this new setting, we introduce PVAD, the first HOI video-3D pairing affordance dataset, providing functional supervision unavailable in prior works. Extensive experiments on PVAD show that VAGNet achieves state-of-the-art performance, significantly outperforming static-based baselines. The code and dataset will be open publicly.

\end{abstract}



%% file: sec/1_introduction_new.tex
\addtocontents{toc}{\protect\setcounter{tocdepth}{0}} 

\section{Introduction}
\label{sec:intro}

\begin{figure*}[t]
  \centering
  \includegraphics[width=\linewidth]{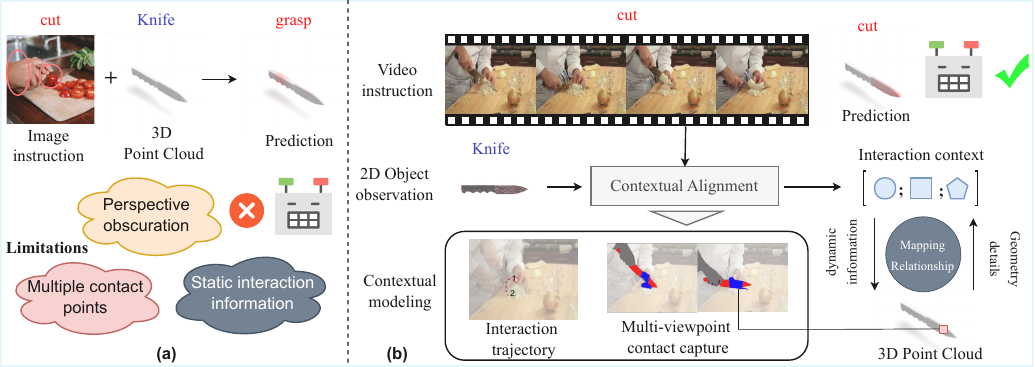}  
  \caption{\textbf{Motivation of Our Work.} (a) Existing 3D affordance grounding methods rely mainly on static visual or textual cues, forcing the model to infer how an object might be used from its shape or static interaction context. Consequently, they struggle with perspective ambiguity, visually similar parts (\eg~blade \vs~handle), and complex multi-contact interactions.
(b) Human-object interaction videos, in contrast, reveal affordance directly through use: they show how hands approach, contact, and move across object surfaces. Our approach leverages this insight by contextually aligning video observations with 3D object geometry and mapping the extracted interaction cues into 3D space. This paradigm shift grounds affordance by observing actual use rather than inferring from appearance, providing richer functional supervision and enabling more reliable 3D affordance grounding.}
  \label{fig:intro}  
\end{figure*}

Humans rarely learn how to use an object merely by observing its shape. Instead, we learn by watching actions, \eg~how a hand approaches and grasps a handle, how motion unfolds, and how contact changes over time (shown in Figure~\ref{fig:intro}). These dynamic cues reveal not only where interaction takes place but also why those regions support specific functions. Thus, affordance is not a static geometric attribute. It is an action-defined relation arising from real human–object interactions (HOI).

3D object affordance grounding aims to localize action-supporting regions on 3D objects~\cite{gibson20133}, enabling downstream embodied tasks such as robotic manipulation~\cite{heidinger20252handedafforder, xu2025a0} and affordance-aware planning~\cite{zhang2025moma}. However, most prior works treat affordance as a purely geometric prediction problem. Given only a point cloud, the model must infer how an object might be used, as though one could understand \textit{“cutting’’} from the silhouette of a knife. This assumption is fragile: geometrically similar parts may serve completely different roles (\eg~a blade versus a handle).

Early efforts rely on geometric features extracted from 3D point clouds~\cite{3daffordancenet, tabib2024lgafford}. Later works incorporate complementary static inputs such as 2D images~\cite{IAGNet, gao2025learning-aaai}, text descriptions~\cite{nguyen2023open}, and 3D Gaussian Splatting~\cite{lu2025geal, wei20253daffordsplat} to provide additional priors. While useful, these modalities remain static and therefore cannot capture the dynamic interaction patterns, \eg~hand trajectories, contact timing, and motion progression that fundamentally define affordance.

Video demonstrations, in contrast, naturally encode how humans \textit{interact with} (use) objects. Prior 2D studies~\cite{fang2018demo2vec, liu2022joint} show that such temporal signals improve affordance localization by highlighting real interaction hotspots. This suggests a key intuition:
\textit{\textbf{To know where interaction occurs, one must observe the interaction.}}

Motivated by this, we introduce video-guided 3D affordance grounding, where the goal is to localize affordance regions on 3D objects by leveraging dynamic HOI videos. This reframes the task from geometry-only inference to motion-conditioned reasoning. The challenge lies in bridging two disparate modalities: videos are dense and sequential, whereas point clouds are sparse and unordered, making spatio-temporal alignment non-trivial.

To address this challenge, we introduce VAGNet (\textbf{V}ideo-guided 3D \textbf{A}ffordance \textbf{G}rounding \textbf{Net}work), which treats interaction video not as an auxiliary signal, but as the primary driver of 3D affordance learning. VAGNet adopts a unified multimodal formulation in which the Multimodal Contextual Alignment Module (MCAM) and the Spatial–Temporal Fusion Module (STFM) operate in concert to jointly couple 3D geometry with evolving interaction dynamics. MCAM establishes cross-modal correspondences between projected 3D object views and video frames through contextual attention, grounding 3D structures in 2D interaction cues.
Subsequently, STFM infuses these interaction-aware representations with frame-level dynamics to capture how contact evolves over time. Through the synergy of MCAM and STFM, VAGNet produces affordance maps that reflect how objects are functionally engaged, rather than merely how they appear.

To support this paradigm, we present \textbf{P}oint \textbf{V}ideo \textbf{A}ffordance \textbf{D}ataset (\textbf{PVAD}), which is the first large-scale dataset pairing HOI videos with 3D object point clouds annotated with affordance regions. PVAD comprises nearly 4K videos and 37K point clouds across 38 object categories and 22 affordance types. 


Our main contributions are summarized as follows:
\begin{itemize}
\item We introduce the task of video-guided 3D object affordance grounding, which leverages HOI videos to provide functional cues for more accurate affordance grounding. This enables affordance to be inferred from \emph{how} objects are actually used rather than only from how they look, addressing the inherent ambiguity of static-based methods.
\item We propose VAGNet, which transforms video-based interaction signals into 3D representations through two dedicated modules. MCAM anchors frame-level interaction evidence onto 3D surfaces to resolve ambiguous or visually similar regions, while STFM injects their temporal evolution to form interaction-aware 3D features. This design enables more reliable localization of functional regions compared to static multimodal baselines.
\item We construct PVAD, the first large-scale HOI video–point-cloud pairing affordance dataset. PVAD provides the necessary supervision for studying this new setting at scale. Experiments on PVAD show that VAGNet outperforms state-of-the-art baselines, demonstrating the value of dynamic interaction signals for precise 3D affordance grounding.

\end{itemize}


%% file: sec/2_related.tex
\section{Related Work}
\label{sec:related}

\textbf{3D Affordance Grounding.} Affordance grounding is a crucial task in robotics. Early research primarily addressed affordance region segmentation in 2D images~\cite{kjellstrom2011visual, koppula2013learning, myers2015affordance, chuang2018learning, do2018affordancenet, li2025analyzing-arxiv-dynamic}. However, as real-world applications often involve 3D environments, recent studies have shifted focus toward affordance grounding in 3D space. Early approaches~\cite{3daffordancenet, tabib2024lgafford} relied solely on isolated point clouds and affordance labels to localize interaction regions. However, these methods often struggle with objects of similar shapes but different functions. To address this limitation, subsequent works have incorporated multimodal inputs as auxiliary guidance. For instance, OpenAD~\cite{nguyen2023open} leverages open-vocabulary text sets to guide 3D affordance grounding. IAGNet~\cite{IAGNet} introduces human-object interaction images as 2D affordance cues, while LMAffordance3D~\cite{zhu2025grounding} jointly learns from both language instructions and interaction images. Building upon IAGNet, GREAT~\cite{great} utilizes large vision-language reasoning models to extract additional textual descriptions of affordance cues from HOI images. More recently, 3DAffordSplat~\cite{wei20253daffordsplat} introduces 3D Gaussian Splatting (3DGS) to capture finer geometric details for affordance reasoning. Despite these advances, existing methods predominantly rely on static information, overlooking the fundamental principle that affordance is defined by dynamic action. As a result, they often fail to identify critical interaction regions and essential geometric details involved in real-world human-object interactions.


\noindent \textbf{Video-Guided 2D Affordance Learning.} Video guidance provides a natural paradigm for imitation learning in robotics, with initial successes demonstrated in 2D domains~\cite{chen2023affordance, liu2022joint, nagarajan2019grounded, bahl2023affordances, li2024learning, heidinger20252handedafforder}. In robotic grasping, Liu \etal~\cite{liu2022joint} proposed a method to generate training data from egocentric videos by localizing interaction regions through hand detection. This strategy has inspired subsequent efforts that further refine techniques on egocentric video data~\cite{zhang2022fine-v5, damen2022rescaling-v1, darkhalil2022epic-v2, grauman2022ego4d-v3, grauman2024ego-v4}. For example, Aff-Grasp~\cite{li2024learning} utilizes pre-contact and contact frames in egocentric video to identify graspable points, which are used to serve as additional inputs for segmentation. In object-specific affordance grounding, several methods employ demonstration videos to guide affordance learning~\cite{fang2018demo2vec, luo2023learning}. Although these video-guided approaches have shown promising results in 2D settings, their potential in 3D affordance domains remains largely underexplored. 


\noindent \textbf{Video-Point Cloud Cross-Modal Fusion.} Cross-modal fusion between videos and point clouds has been explored in several related domains. For talking head synthesis, PointTalk~\cite{xie2025pointtalk} leverages 3DGS to convert reference videos into a tri-plane representation, which is then synchronized and fused with dynamic lip point cloud features via an attention mechanism. In 3D scene understanding, Video-3D LLM~\cite{zheng2025video} integrates scene demonstration videos with 3D scenes using a Video LLM~\cite{zhang2024video-videollm1, qwen2technicalreport-videollm2}. However, this method requires pre-rendering 3D point clouds into 2D RGB images, leading to an inevitable loss of structural information. More related to our task, Ego-SAG~\cite{liu2024grounding-ego3d} introduced a task for grounding 3D scene affordance from egocentric videos using learnable bilateral queries. However, the egocentric view often lacks sufficient scene context in complex environments, and their network is resource-intensive. In contrast, our approach first aligns objects with video sequences and models contextual relationships in the 2D space, then injects this information into point cloud features through learned mapping relations, resulting in a more lightweight and efficient modality fusion.



%% file: sec/3_method.tex
\section{Method}
\label{sec:method}

\begin{figure*}[t]
  \centering
  \includegraphics[width=\linewidth]{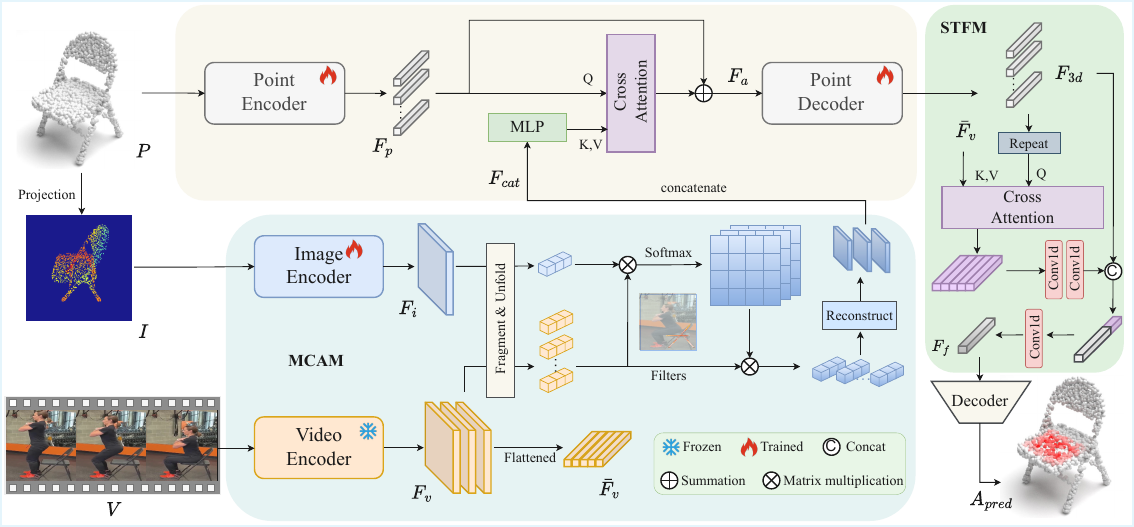}  
  \caption{\textbf{VAGNet Architecture.} Our model takes a point cloud, its 2D projection, and a corresponding interaction video as input. These
  are first processed by three modality-specific encoders to extract point features ($F_p$), image features ($F_i$), and video features ($F_v$). Then, the Multimodal Contextual Alignment Module (MCAM) aligns $F_i$ with $F_v$ to produce a joint 2D representation $F_{2d}$, which is then fused with $F_p$ through a cross-attention layer and a point decoder to obtain the context-aligned 3D feature $F_{3d}$. Subsequently, the Spatial-Temporal Fusion Module (STFM) integrates $F_v$ and $F_{3d}$ to produce the spatio-temporal feature $F_f$, which is finally fed into a decoder to generate the 3D affordance mask.}
  \label{fig:pipeline}  
\end{figure*}

\subsection{Overview}
\label{subsec:overview}

Given a 3D object represented as a point cloud $P \in \mathbb{R}^{N \times 3}$ with $N$ points and a corresponding interaction video $V \in \mathbb{R}^{T \times 3 \times H \times W}$, our goal is to predict its affordance mask $A \in \mathbb{R}^{N \times 1}$ through an end-to-end model $f$, such that $A=f(P,V)$. As illustrated in Figure~\ref{fig:pipeline}, our proposed VAGNet processes the inputs in several stages. First, the point cloud is projected to a 2D plane using affordance-aware camera parameters to generate an auxiliary object view. The point cloud $P$, its projection $I$, and the video $V$ are then encoded by PointNet++ \cite{qi2017pointnet++}, ResNet \cite{he2016deep-resnet}, and TimeSformer \cite{bertasius2021space-timesformer} to produce features $F_p \in \mathbb{R}^{C \times N_p}$, $F_i \in \mathbb{R}^{C \times H' \times W'}$, and $F_v \in \mathbb{R}^{T \times C \times H' \times W'}$, respectively. The Multimodal Contextual Alignment Module (MCAM) then aligns $F_i$ and $F_v$ in the 2D space, yielding a joint 2D representation $F_{2d}$ that encapsulates video-derived interaction contexts. This is followed by a cross-attention mechanism that fuses the point cloud feature $F_p$ with $F_{2d}$ in the 3D space, effectively injecting the 2D contextual cues into the 3D representation to form $F_{3d}$ (Sec.~\ref{sec: mcam}). Finally, the Spatial-Temporal Fusion Module (STFM) integrates the context-enhanced 3D feature $F_{3d}$ with the temporal video feature $F_v$ (Sec.~\ref{sec: stfm}), producing a spatio-temporal feature $F_f$ that is decoded to obtain the final affordance mask (Sec.~\ref{sec: decode}). To facilitate this research, we first construct a dataset (PVAD), which contains paired HOI videos and 3D object point clouds (Sec.~\ref{sec: datasets}).

\subsection{Point Cloud-Video Pairing (PVAD) Dataset}
\label{sec: datasets}

A significant barrier in video-guided 3D affordance grounding has been the lack of a dedicated dataset pairing HOI videos with 3D object point clouds. To address this, we introduce the \textbf{P}oint-\textbf{V}ideo \textbf{A}ffordance \textbf{D}ataset \textbf{(PVAD)}, a pioneering, large-scale benchmark. The main challenge in constructing such a dataset lies in collecting HOI videos where the objects exhibit geometric consistency with their corresponding point cloud models. PVAD is carefully curated, containing 3,763 interaction videos and 36,765 point clouds spanning 38 object categories and 22 affordance types, with an overview provided in Figure~\ref{fig:dataset} (see Appendix for details).

The video data in PVAD are primarily collected from human action recognition datasets, including UCF101~\cite{soomro2012ucf101}, HMDB~\cite{kuehne2011hmdb}, supplemented with clips from other public sources. The corresponding 3D point clouds are mainly sourced from 3DAffordanceNet~\cite{3daffordancenet} and PIADv2~\cite{great}. All video clips undergo a preprocessing stage to ensure that each of them unambiguously depicts a complete human-object interaction and is trimmed to a maximum duration of 10 seconds. Following prior works~\cite{li2024laso, gao2025learning-aaai}, we divide PVAD into two evaluation settings: \textbf{Seen} and \textbf{Unseen}. In the \textbf{Seen} setting, the object-affordance pairing patterns are shared between the training and evaluation phases. Conversely, in the \textbf{Unseen} setting, the pairings in evaluation are distinct from those used in training.

\subsection{2D-3D Alignment with Contextual Attention}
\label{sec: mcam}

To bridge the modality gap between static 3D point clouds and dynamic 2D videos, we employ projection as a direct alignment mechanism. This module begins by generating an adaptive-viewpoint projection of the 3D object, which is subsequently aligned with video frames to model contextual information in 2D space. The camera parameters for this projection are optimized using a view planning approach~\cite{zeng2020view-camera} to actively seek viewpoints that are rich in dynamic interaction cues relevant to affordance.

While the projected image contains only the object, the video frames encompass the object, human, and environment. To align the isolated object in the projection with its contextual scene in the video frames, we introduce a contextual attention mechanism. Here, the projection feature $F_i$ serves as the foreground, and each video frame feature in $F_v = \{f_v^{(t)}\}_{t=1}^T$ provides the background context, forming $T$ contextual attention pairs (i.e., foreground-background pairs). Both foreground and background features are partitioned into multiple $3 \times 3$ patches with a specified stride, and flattened into patch features $f$ and $b_t$. The normalized inner product between them is computed:
\begin{equation}
    A_t = \text{softmax}\left(\langle\frac{f}{\|f\|},\frac{ b_t}{\| b_t\|}\rangle \cdot d^{-\frac{1}{2}}\right).
\end{equation}
where each value in $A_t \in \mathbb{R}^{  L \times L}$ denotes the similarity between a foreground patch and a background patch, $L=H' \times W'$ is the sequence length of patches, and $d=3 \times 3 \times C$ is the channel dimension for scaling. The background patches are then reused as deconvolutional filters to reconstruct the foreground: 
\begin{equation}
    F_i^{(t)} = \Gamma(A_t \cdot b_t),
\end{equation}
where $\Gamma(\cdot)$ represents a folding operation that reassembles the sliding patches back to a feature map, and $F_i^{(t)} \in \mathbb{R}^{  C \times H' \times W'}$ denotes the reconstructed projected image. Due to the significant content discrepancy between the object projection (foreground) and video frames (background), we omit the attention propagation step used in standard contextual attention~\cite{yu2018generative-contextual}.

\begin{figure*}[t]
  \centering
  \includegraphics[width=\linewidth]{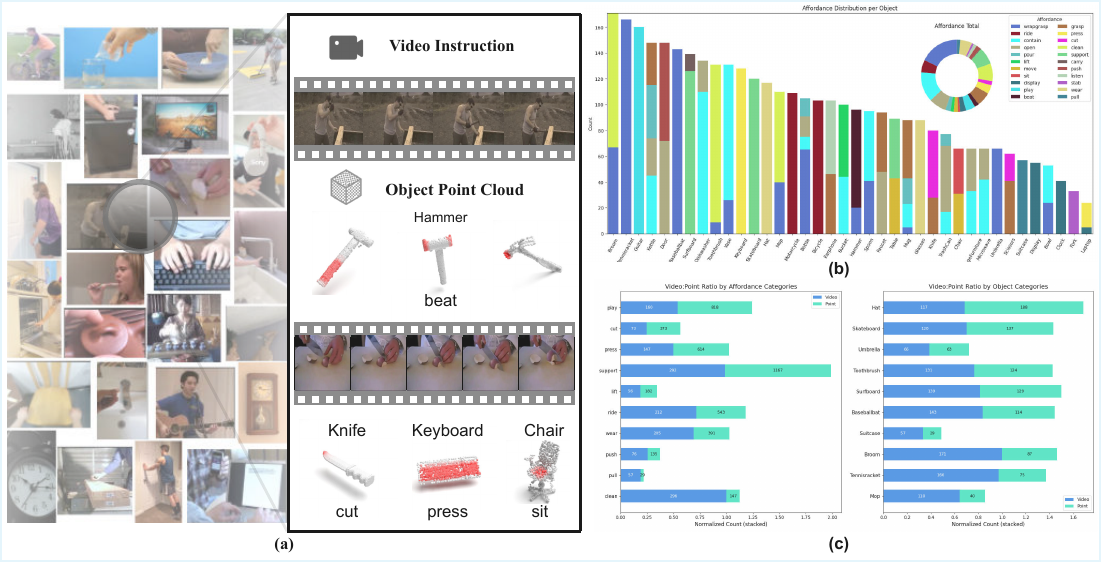}  
  \caption{\textbf{The PVAD Dataset.} (a) Overview with annotated affordance regions on point clouds highlighted in red. (b) Distribution of video samples across different affordance categories. (c) Statistics of video and point cloud counts for representative object-affordance pairs. 
  } 
  \label{fig:dataset}
\end{figure*}

After computing the contextual attention between the projected image and each video frame, we concatenate the resulting frame-wise contextual features $\{F_i^{(1)}, F_i^{(2)}, \cdots, F_i^{(T)}\}$ along the temporal dimension to form $F_{cat}\in \mathbb{R}^{  T \times C \times H' \times W'}$. To generate the information of dynamic variation from different frames, we apply an Multi-Layer Perceptron (MLP) to perform temporal fusion on the concatenated features, aggregating them into a unified 2D representation: 
\begin{equation}
    F_{2d} = \phi(\text{BN}(W_2 \cdot \phi(W_1 \cdot F_{cat}))),
\end{equation}
where $\phi(\cdot)$ is the ReLU activation, $\text{BN}(\cdot)$ is batch normalization, and the matrices $W_1$ and $W_2$ are learnable weights for temporal compression.

Upon obtaining the feature $F_{2d}\in \mathbb{R}^{  C \times H' \times W'}$ enriched with contextual information from the video, we leverage the geometric mapping between the point cloud and its projection to transfer the dynamic interaction-based affordance features back to the 3D feature space. Specifically, the point cloud feature $F_{p}$ performs a linear transformation to generate the query $Q=W_qF_p$, while the flattened $F_{2d}$ is linearly projected to obtain the key $K=W_kF_{2d}$ and value $V=W_vF_{2d}$, where $W_q, W_k,W_v$ are projection matrices. Subsequently, a cross-attention mechanism is employed to inject the 2D context-aware feature into the 3D point cloud feature to obtain the augmented feature: 
\begin{equation}
    F_{a} = F_p + (\text{softmax}(Q^T\cdot K / \sqrt{d'})\cdot V^T)^T,
\end{equation}
where $Q \in \mathbb{R}^{N_p \times C}$, $K,V \in \mathbb{R}^{(H' \cdot W') \times C}$, $F_{a} \in \mathbb{R}^{C \times N_p}$, and $d'$ is the channel dimension for scaling. Afterwards, we upsample the $F_a$ to obtain the context-aligned 3D branch feature $F_{3d} \in \mathbb{R}^{C \times N}$ using the decoder of PointNet++~\cite{qi2017pointnet++}.


\subsection{Spatial-Temporal Fusion Module}
\label{sec: stfm}

The preceding MCAM injects the 2D contextual interaction information into the point cloud feature, enabling VAGNet to effectively model temporal dynamics and global geometric correspondences. To further comprehensively integrate interaction cues across both spatial and temporal dimensions, we design a Spatial–Temporal Fusion Module (STFM). As shown in Figure~\ref{fig:pipeline}, this module enhances the context-aligned 3D branch feature $F_{3d}$ by progressively fusing it with the temporally-aware video feature $F_v$, enabling the model to understand how interactions evolve in 3D space over time.

Specifically, we first establish a frame-wise correspondence by repeating $F_{3d}$ across all $T$ time steps of video frames, resulting in $\bar{F}_{3d} \in \mathbb{R}^{ T \times C \times N}$. The video feature $F_v$ is flattened into a sequence of patches and arranged in chronological order of frames, which are then reshaped to $\bar{F}_v \in \mathbb{R}^{T \times C \times L}$. A cross-attention operation is then applied between the 3D feature and the video feature: 
\begin{equation}
    F_{pv} = \Theta(\text{CrossAttn}(\bar{F}_{3d}, \bar{F}_v)),
\end{equation}
where $\Theta(\cdot)$ denotes $1 \times 1$ convolutional layers along the time dimension and $F_{pv} \in \mathbb{R}^{  C \times N}$. This operation allows the geometrically-grounded 3D features to attend to dynamic visual contexts, modeling the object's interaction process in 3D space.

Finally, we inject the dynamic information into the context-aligned 3D branch feature to produce the spatio-temporal feature
\begin{equation}
    F_f = \Theta([F_{3d}, F_{pv}]),
\end{equation}
where $[\cdot]$ denotes concatenation along the channel dimension, and $F_f \in \mathbb{R}^{  C \times N}$. This allows each 3D point to attend to its corresponding visual context in the temporal sequence, capturing critical interaction-dependent cues. 

\begin{table*}[t]
\caption{\textbf{Comparison on the PVAD Dataset}. The best results are marked in \textbf{bold}. The second-best results are \underline{underlined}. The dataset is divided into two partitions: \textbf{Seen} and \textbf{Unseen}. AUC and aIOU are shown in percentage.}
\label{tab:pvad}
\centering
\setlength{\tabcolsep}{10pt}
\begin{tabular}{c|cccc|cccc}
\toprule
 & \multicolumn{4}{c}{\textbf{Seen}} & \multicolumn{4}{|c}{\textbf{Unseen}} \\
\midrule
\textbf{Method} & \textbf{AUC ↑} & \textbf{aIOU ↑} & \textbf{SIM ↑} & \textbf{MAE ↓} & 
\textbf{AUC ↑} & \textbf{aIOU ↑} & \textbf{SIM ↑} & \textbf{MAE ↓} \\
\midrule
\multicolumn{9}{c}{Image-3D Alignment Methods} \\
\midrule
XMF~\cite{aiello2022cross-xmf} & 90.83& 35.85& 0.668& 0.074& 57.22& 8.05& 0.268& 0.315 \\
IAGNet~\cite{IAGNet} & 93.04& 39.77& 0.682& 0.072& 58.91& 10.19& 0.297& 0.305 \\
GREAT~\cite{great} & \underline{93.75}& 40.23& 0.703& 0.066& 59.83& 10.42& \underline{0.302}& 0.265 \\
VAGNet-img (ours) & 93.74& \underline{41.58} & \underline{0.707} & \underline{0.065} & 
\underline{60.22} & \underline{11.03} & 0.287& \underline{0.248} \\
\midrule
\multicolumn{9}{c}{Video-3D Alignment Methods} \\
\midrule
Baseline & 89.35& 34.15& 0.604& 0.096& 55.38& 7.71& 0.254& 0.322 \\
VAGNet (ours) & \textbf{94.33}& \textbf{42.96}& \textbf{0.723}& \textbf{0.061}& 
\textbf{61.31}& \textbf{12.09}& \textbf{0.304}& \textbf{0.201} \\
\bottomrule
\end{tabular}
\end{table*}

\subsection{Affordance Decoding and Loss Function}
\label{sec: decode}

The output of STFM, the spatio-temporal feature $F_f$, encapsulates fine-grained affordance localization by combining geometric precision from 3D structures and dynamic contextual information from video motion. To generate the final affordance map, we employ a lightweight affordance decoder, which takes the spatio-temporal feature $F_f$ as input and produces the affordance map at the point level:
\begin{equation}
    A_{\text{pred}} = \sigma(\text{MLP}(F_f)),
\end{equation}
where $\sigma(\cdot)$ is the sigmoid function, $\text{MLP}(\cdot)$ denotes the Multi-Layer Perceptron, and $A_{\text{pred}} \in \mathbb{R}^{N \times 1}$. 

The entire framework of VAGNet is trained end-to-end by optimizing a combined loss function designed for robust segmentation performance. The total loss $\mathcal{L}_{\text{total}}$ is defined as the sum of Focal loss~\cite{lin2017focal} and Dice loss~\cite{milletari2016v_dice},
supervising the point-wise heatmap on 3D objects' point clouds: 
\begin{equation}
    \mathcal{L}_{\text{total}} = \mathcal{L}_{\text{focal}} + \mathcal{L}_{\text{dice}}.
\end{equation}

%% file: sec/4_experiments.tex

\section{Experiments}
\label{sec:experiments}

\subsection{Setup}

\textbf{Implementation Details.} We utilize PointNet++~\cite{qi2017pointnet++} as the 3D backbone. The image encoder is built with a ResNet18~\cite{he2016deep-resnet} as the 2D backbone. The video branch employs TimeSformer~\cite{bertasius2021space-timesformer} pre-trained on Kinetics-600~\cite{kay2017kinetics-600} as the video encoder to leverage its human motion priors, with parameters frozen during training. We uniformly sample 8 frames per video. We train VAGNet for 60 epochs with batch size 12 on a single NVIDIA RTX 4090 GPU, using the AdamW optimizer~\cite{loshchilov2017decoupled-adam} with a cosine learning rate scheduler~\cite{loshchilov2016sgdr-cosine}. The model is trained with an initial learning rate of 1e-4 and a weight decay to 1e-6.

\noindent\textbf{Baselines.} Since no existing 3D affordance grounding work utilizes both video and point cloud as inputs, we choose state-of-the-art image-3D alignment methods in 3D affordance grounding as baselines for comprehensive comparison. The most relevant works include IAGNet~\cite{IAGNet} and GREAT~\cite{great}, both of which leverage HOI image as additional input to guide affordance learning. Besides, we include the image-point cloud fusion framework XMF~\cite{aiello2022cross-xmf} and adapt it for affordance mask generation. Furthermore, we also adapt our method (VAGNet-img) to use the same inputs (image and point cloud) for a more fair comparison. Specifically, the input image is replicated and structured into a sequence of $T$ frames to match the input format required by VAGNet. For those image-3D alignment methods, we choose one HOI frame in the video sample as the input image, following the setup of PIAD~\cite{IAGNet} and PIADv2~\cite{great}. Additionally, we incorporates the video-point cloud fusion module from PointTalk~\cite{xie2025pointtalk} to construct a video-3D alignment Baseline, enabling more complete comparison with our dataset. Please refer to the appendix for more details.

\noindent\textbf{Evaluation Metrics.} Consistent with previous works~\cite{li2024laso, lu2025geal, chu20253d-LLM}, we evaluate performance using four established metrics: \textbf{AUC}~\cite{lobo2008auc}, \textbf{aIoU}~\cite{rahman2016optimizing-aiou}, \textbf{SIM}~\cite{swain1991color-sim}, and \textbf{MAE}~\cite{willmott2005advantages-mae}.


\subsection{Comparison Results}

\begin{figure*}[t]
  \centering
  \includegraphics[width=\linewidth]{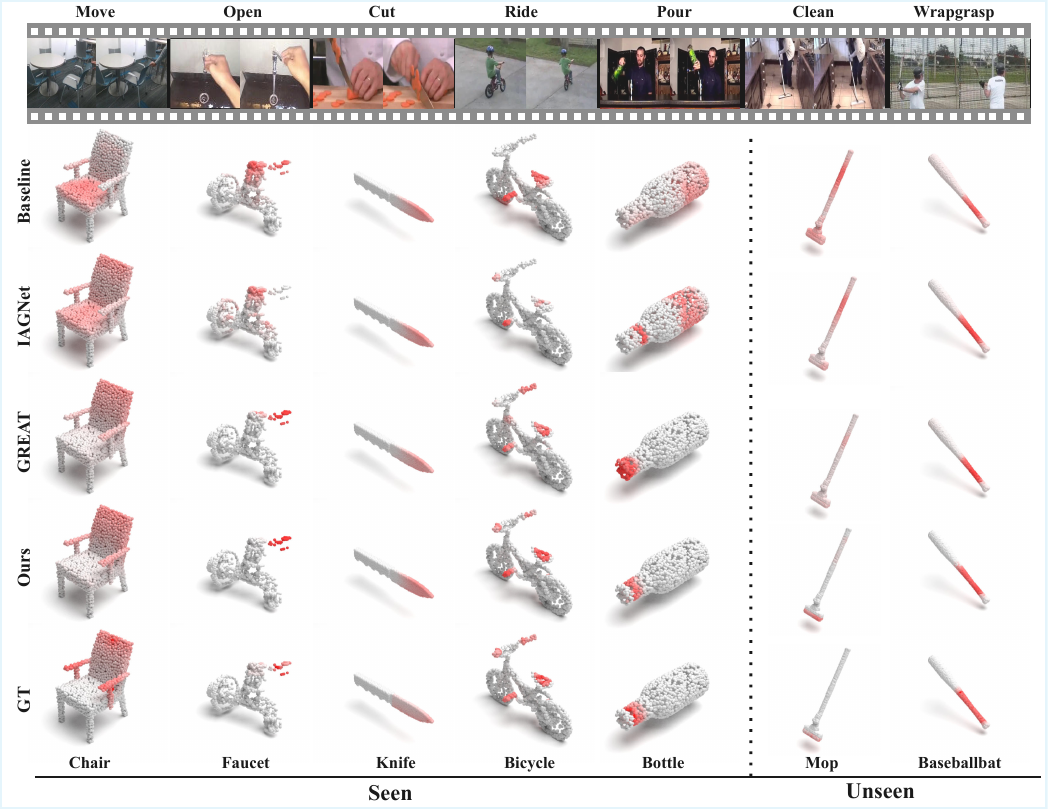}  
  \caption{\textbf{Comparative visualization} of affordance results generated by different methods, on selected test instances from \textbf{Seen} and \textbf{Unseen} settings. Each object is associated with its corresponding interaction video, represented by two sampled frames. The affordance probability for each 3D point is encoded in a heatmap (red means high probability). Ground truth (GT) are provided for reference.} 
  \label{fig:vis-1}
\end{figure*}

\textbf{Quantitative Analysis.} We compare the performance of VAGNet-img and VAGNet against recent image-3D alignment methods and aforementioned baselines on our PVAD dataset. The results are summarized in Table~\ref{tab:pvad}. First, even under identical input conditions (image and point cloud), our VAGNet-img outperforms existing image–3D alignment methods in 3D affordance grounding across most metrics, demonstrating the effectiveness and robustness of our approach. More importantly, when leveraging full video sequences, VAGNet achieves state-of-the-art performance in both \textbf{Seen} and \textbf{Unseen} settings, significantly surpassing all baselines. Specifically, under the \textbf{Seen} setting, VAGNet outperforms the strongest baseline GREAT by notable margins of +2.73 in \textbf{aIoU} and +0.02 in \textbf{SIM}. In the more challenging \textbf{Unseen} setting, our method maintains superior generalization, exceeding GREAT by +1.48 in \textbf{AUC} and +1.67 in \textbf{aIoU}. We also observe that both the shape completion method XMF~\cite{aiello2022cross-xmf} and the video-3D baseline adapted from PointTalk~\cite{xie2025pointtalk} underperform compared to dedicated affordance grounding methods, highlighting the specificity of our task.


\noindent\textbf{Qualitative Analysis.} Figure~\ref{fig:vis-1} illustrates visualization results of video-guided 3D affordance grounding under \textbf{Seen} and \textbf{Unseen} settings. VAGNet produces more accurate and complete results that align closely with the ground truth, whereas other methods exhibit noticeable shortcomings. For example, when grounding the ``ride'' affordance on a bicycle, both IAGNet and GREAT fail to cover all relevant parts due to their reliance on a single, static image perspective. In contrast, VAGNet successfully identifies the entire functional region by effectively integrating dynamic interaction patterns from the video, demonstrating the critical advantage of leveraging temporal cues for precise 3D affordance grounding.

\subsection{Ablation Studies}

\begin{figure}[t]
    \centering
    \includegraphics[width=\linewidth]{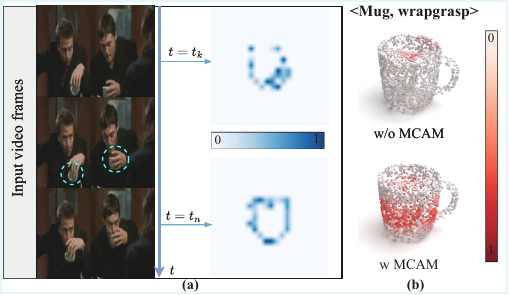}
    \caption{\textbf{Ablation of MCAM.} (a) Contextual Attention Maps during the alignment between project image and video instruction. (b) Visualization results with/without MCAM.}
    \label{fig:contextual}
\end{figure}

We evaluate the impact of our core architectural components. Table~\ref{tab:ablation} shows the effects of MCAM, STFM and the whole 2D branch under \textbf{Seen} and \textbf{Unseen} settings. First, we remove the entire 2D branch and MCAM to evaluate the effect of directly fusing point cloud feature with video feature through STFM. This setting results in the most substantial performance decline across all metrics, since the STFM relies on the context-aligned 3D feature provided by the MCAM for effective fusion. Second, we remove STFM and directly pass the output of 3D branch to the affordance decoder, which also leads to a notable performance degradation compared to the full model. Finally, we perform an ablation on the MCAM module by dropping the contextual attention module and directly fusing the projection feature with 3D branch. The results demonstrate a performance improvement over the 2D branch ablation. As shown in Figure~\ref{fig:contextual} (a), we visualize the attention maps from MCAM for a representative example. As the projected image progressively integrate with interaction frames, MCAM effectively captures affordance-relevant regions.

\begin{table}[t]
\caption{\textbf{Abalation Studies.} We evaluate the contribution of each key module: the multimodal contextual alignment module \textbf{MCAM}, the spatial-temporal fusion module \textbf{STFM}, and the entire 2D branch (\textbf{Proj}). AUC and aIOU are shown in percentage. Best results are in \textbf{bold}.}
\label{tab:ablation}
\centering
\resizebox{\linewidth}{!}{
    \begin{tabular}{c|ccc|cccc}
    \toprule
    \textbf{Type} & \textbf{MCAM} & \textbf{STFM} & \textbf{Proj} 
    & \textbf{AUC ↑} & \textbf{aIOU ↑} & \textbf{SIM ↑} & \textbf{MAE ↓} \\
    \midrule
    \multirow{4}{*}{\textbf{Seen}} 
    & &\textbf{\checkmark}& & 92.97& 39.94& 0.682& 0.072 \\
    &\textbf{\checkmark}&  &\textbf{\checkmark} & 93.83& 39.78& 0.694& 0.066 \\
    & &\textbf{\checkmark}&\textbf{\checkmark}& 93.87& 41.86& 0.711& 0.064 \\
    & \textbf{\checkmark} & \textbf{\checkmark} & \textbf{\checkmark} & 
    \textbf{94.33}& \textbf{42.96}& \textbf{0.723}& \textbf{0.061} \\
    \midrule
    \multirow{4}{*}{\textbf{Unseen}} 
    & &\textbf{\checkmark}& & 59.26& 11.64& 0.288& 0.279 \\
    &\textbf{\checkmark}&  &\textbf{\checkmark} & 59.88& 10.25& 0.302& 0.304 \\
    & &\textbf{\checkmark}&\textbf{\checkmark}& 60.89& 11.98& 0.297& 0.263 \\
    & \textbf{\checkmark} & \textbf{\checkmark} & \textbf{\checkmark} & 
    \textbf{61.31}& \textbf{12.09}& \textbf{0.304}& \textbf{0.201} \\
    \bottomrule
    \end{tabular}
}
\end{table}

\subsection{Performance Analysis}

\begin{figure}[t]
  \centering
  \includegraphics[width=\linewidth]{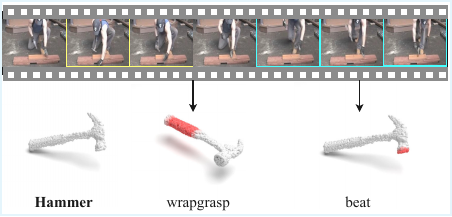}  
  \caption{\textbf{Single Instruction with Multiple Affordances.} A single video may capture different affordances of the same object, and the model should localize the most relevant one (``beat'').}
  \label{fig:multiafd}
\end{figure}

\textbf{Single Instruction with Multiple Affordances.} A single video often contains multiple affordances with the same object. As shown in Figure~\ref{fig:multiafd}, the early and later parts of the video present different affordances of a ``hammer'', and our method can focus on the key clip of ``beat'' affordance in the later part. Thus, given the same video instruction, VAGNet demonstrates a strong capability to identify and localize the most prominent affordance regions in the current clip.

\begin{figure}[t]
  \centering
  \includegraphics[width=\linewidth]{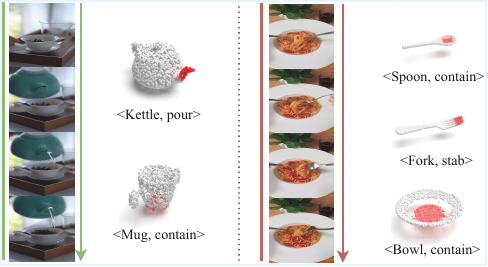}  
  \caption{\textbf{Single Instruction with Multiple Objects.} Grounding affordances on different objects with the same video that contains multiple interactions.}
  \label{fig:multiobj}
\end{figure}

\noindent\textbf{Single Instruction with Multiple Objects.} A single video can also involve multiple objects and sometimes encompasses several distinct interactions. As shown in Figure~\ref{fig:multiobj} (left), pouring water from a kettle to a mug entails two separate object-affordance pairs. When taking the same video instruction along with different objects as inputs, our model can accurately localize affordance regions relevant to each specific object without being confounded by the presence of other interacting objects in the scene.


%% file: sec/5_conclusion.tex
\section{Conclusion}
\label{sec:conclu}

In this paper, we introduce video-guided 3D object affordance grounding, a novel task that leverages dynamic HOI videos to localize functional regions on 3D objects. To address this, we propose VAGNet, a framework that aligns 3D object geometry with contextual video cues and models dynamic affordance processes in 3D space, leading to more accurate and context-aware grounding results. Moreover, we construct PVAD, the first large-scale dataset pairing HOI videos with corresponding 3D object point clouds, establishing a benchmark for this task. Extensive experiments on PVAD demonstrate that VAGNet achieves superior performance compared to existing methods, confirming the effectiveness of incorporating dynamic video guidance for 3D affordance learning.

This work opens several promising research directions. First, extending video guidance to interactive 4D scenes could enable temporal reasoning over continuously evolving environments. Second, incorporating language supervision (\eg~action verbs or natural-language descriptions) with video cues may further enhance cross-modal understanding. Third, designing scalable and more efficient architectures for large-scale 3D–video fusion could facilitate deployment in real-time robotic systems. We believe these avenues will significantly advance the integration of perceptual learning and embodied intelligence.

\vspace{\baselineskip}